\documentclass[conference]{IEEEtran}
\usepackage{cite,amsmath,amssymb,amsfonts,graphicx,textcomp,xcolor}
\usepackage{booktabs,balance,tikz}
\def\BibTeX{{\rm B\kern-.05em{\sc i\kern-.025em b}\kern-.08em
    T\kern-.1667em\lower.7ex\hbox{E}\kern-.125emX}}
\usetikzlibrary{arrows.meta}

\newcommand{\RevisionStatus}{}
\newcommand{\ConclusionResults}{DEQ-Det combines device morphology and layout context for few-shot analog IC detection without additional background annotations. Support examples update task-specific references, allowing new target geometries to be specified without separate labels for dummy or routing structures. Macro AP50 exceeds the strongest listed baselines by \TenShotGain{} and \ThirtyShotGain{} points at 10 and 30 shots. Matched controls show gains from support-guided query allocation and quality-aware ranking. RoI background evidence reduces false positives at comparable recall, yielding fewer distracting candidates for manual review. With $K$ annotated devices defining the inspection target, the predicted boxes can guide review toward localized regions in electron micrographs. These locations also provide spatial references for circuit analysis and selection of probing regions.}
\newcommand{\TenShotGain}{3.97}
\newcommand{\ThirtyShotGain}{9.71}

\begin{document}
\title{DEQ-Det: Dual Evidence Queries for Few-Shot Analog IC Device Detection}
\author{
\IEEEauthorblockN{Zidong Zhao\textsuperscript{1}, Rong Huang\textsuperscript{2}, Chengxing Fan\textsuperscript{1}, Jie Hu\textsuperscript{1},\\
Malik Haris\textsuperscript{1}, Ghazal A. Fahmy\textsuperscript{3}, Morteza Mousazadeh\textsuperscript{4}, Guoqiang Zhang\textsuperscript{5\textdagger}}
\IEEEauthorblockA{\textsuperscript{1}Zhejiang University, China, \textsuperscript{2}Donghua University, China,\\
\textsuperscript{3}National Telecommunication Institute, Egypt, \textsuperscript{4}Urmia University, Iran,\\
\textsuperscript{5}Ningbo Tech University, China}}
\maketitle
\begingroup
\renewcommand{\thefootnote}{\fnsymbol{footnote}}
\footnotetext[2]{Corresponding author: Guoqiang Zhang (guoqiang\_zhang@nbt.edu.cn).}
\endgroup
\suppressfloats[t]

\begin{abstract}
Detecting analog integrated circuit (IC) devices in layout images supports circuit analysis and inspection. With only a few annotated examples, variation in device geometry and visual similarity to dummy or routing structures make reliable detection difficult. Foreground similarity alone can highlight plausible targets without distinguishing them from competing layout patterns. We propose Dual Evidence Queries for Detection (DEQ-Det), which constructs complementary foreground and background references from the same support images without additional annotations. Foreground references capture target morphology, while selected background references characterize surrounding layout patterns. Both sources guide query selection and region verification, combining target appearance with layout context. Across five held-out device categories and three random seeds, DEQ-Det achieves macro AP50 of 39.41 and 53.03 at 10 and 30 shots, exceeding the strongest listed baselines by \TenShotGain{} and \ThirtyShotGain{} points, respectively. A matched 30-shot ablation further shows that RoI background evidence reduces pooled false positives at comparable recall, supporting its role in interference suppression.
\end{abstract}
\RevisionStatus

\section{Introduction}
Integrated circuit (IC) device detection localizes resistors, capacitors, and transistors in layout images for circuit analysis, targeted inspection, and physical probing~\cite{ravikumar2022laserprobing}. Layout representations can be reconstructed from optical microscopy or scanning electron microscopy (SEM) images~\cite{shao2021iclayout}. Analog devices have customized geometries reflecting sizing and matching requirements~\cite{lienig2020layout}, limiting the transferability of digital standard-cell templates to analog layouts. Dummy fills and repetitive routing patterns can resemble devices and cause false positives. Automated localization can focus manual inspection on candidate regions. Dense annotation for supervised detection is costly, with additional labeling often needed across fabrication processes.

Few-shot object detection (FSOD) adapts to targets defined by a small annotated support set~\cite{kang2019fewshot,wang2020tfa}. Existing approaches improve target representation through support-conditioned aggregation, contrastive learning, and prototype distillation~\cite{yan2019meta,sun2021fsce,han2023vfa,wang2024fpd}. DE-ViT and FSOD-VFM use transferable features for exemplar-based or cross-domain detection~\cite{zhang2025devit,feng2026fsodvfm}. For IC imagery, SSRNet combines few-shot segmentation and a region classifier for packaged ICs in optical PCB images~\cite{wang2025ssrnet}. Detecting internal devices in analog layouts requires resolving customized geometries amid repeated, target-like structures.

\begin{figure}[t]
\centering
\includegraphics[pagebox=cropbox,width=\columnwidth]{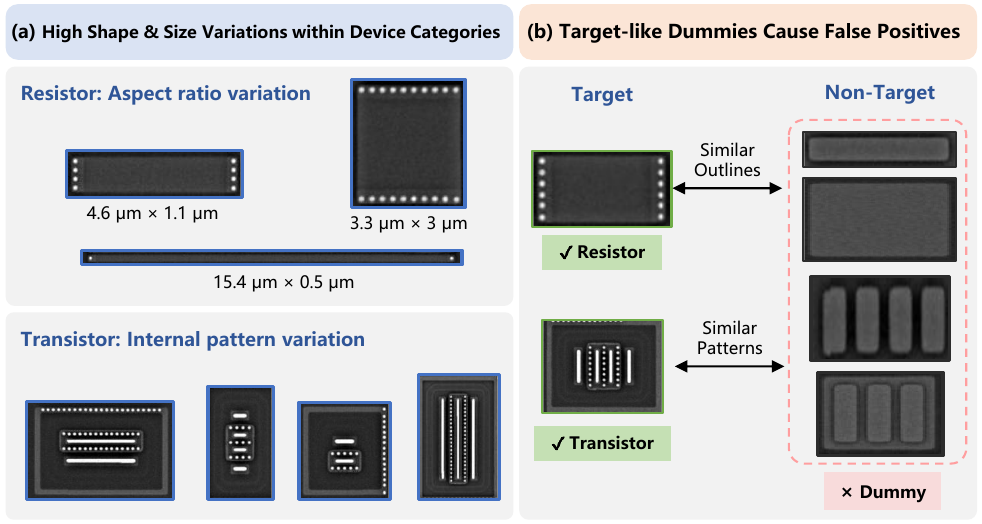}
\caption{Analog layout variability and ambiguity. Device geometries vary within categories; non-target dummy structures share outlines or repeated patterns with devices. Blue/green boxes indicate devices, and the red dashed region indicates dummies.}
\label{fig:motivation}
\end{figure}

\begin{figure*}[!t]
\centering
\includegraphics[pagebox=cropbox,width=0.74\textwidth]{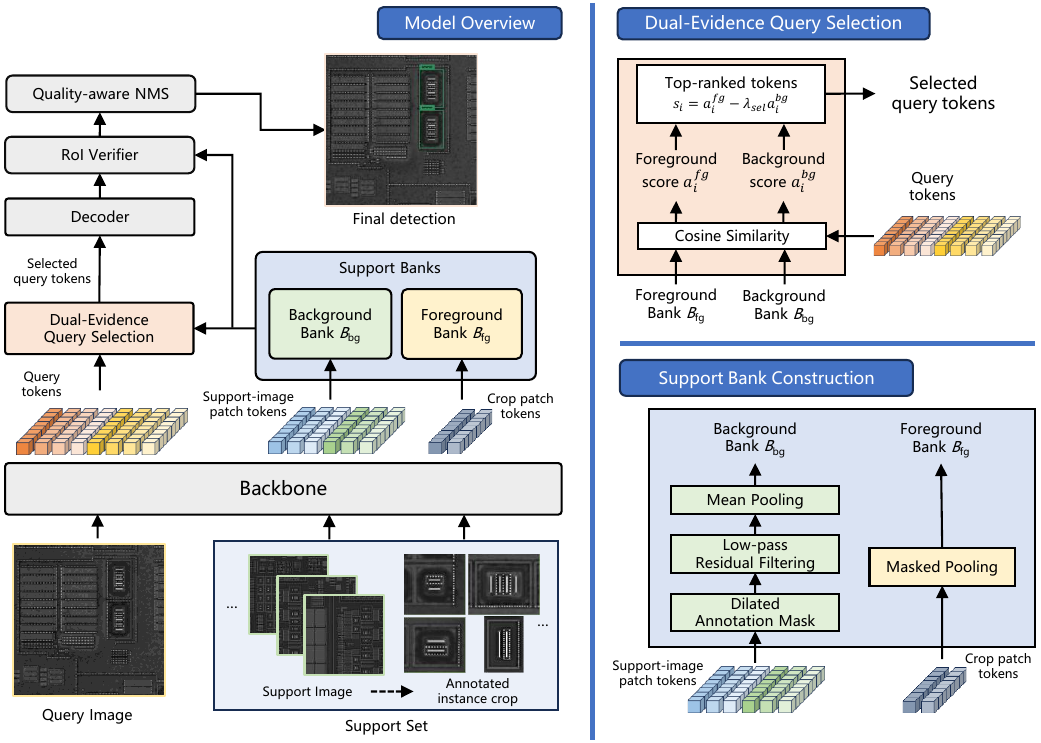}
\caption{DEQ-Det. Foreground references guide target selection, and filtered background references supply layout context for interference suppression. Their means guide query selection, while individual bank records enter the RoI relation head. Foreground probability and predicted localization quality jointly rank candidate detections.}
\label{fig:pipeline}
\end{figure*}

Figure~\ref{fig:motivation} illustrates device variability and resemblance to non-target structures. Foreground similarity can highlight plausible candidates while also responding to dummy fills or routing patterns. The same support images provide context that helps resolve this ambiguity: annotated devices specify the requested morphology, while selected background regions characterize competing layout patterns. Using both sources allows candidate selection and verification to account for the target and its surroundings.

We propose \textbf{Dual Evidence Queries for Detection (DEQ-Det)} to integrate device morphology and support-specific layout context throughout detection. Our contributions are twofold: (1) a dual-evidence framework that derives complementary foreground and background references without additional annotations and uses them in query selection and region verification to couple target selection with interference suppression; and (2) evaluation across five held-out device categories and three random seeds, demonstrating system-level AP50 gains. Matched controls establish the roles of support-guided allocation and quality-aware ranking, while a RoI background control shows fewer pooled false positives at \mbox{comparable recall}.

\section{Proposed Method}
\subsection{Support-Derived Foreground and Background}
Given $K$ selected boxes of one novel category, DEQ-Det constructs two evidence banks (Fig.~\ref{fig:pipeline}). A shared DINOv2 ViT-L/14 encoder~\cite{oquab2024dinov2}, with a frozen IC reconstruction adapter, produces 1,024-dimensional features. Each support box is enlarged by 20\%, cropped, resized while preserving its aspect ratio, and padded to $336\times336$ before encoding in its original orientation. Masked pooling over the annotated region yields one morphology record per instance, giving $B_{fg}=\{f_k\}_{k=1}^{K}$.

Background extraction summarizes the surroundings of annotated devices in full support images, using spatial exclusion and residual filtering to favor stable layout interiors. Valid feature patches outside a two-cell dilation of the selected boxes form a candidate mask $C_s$. The channel-averaged absolute difference between the image and its Gaussian-blurred version ($15\times15$ kernel, $\sigma=3$) measures local texture variation, providing a visual heuristic for background selection. Quantiles are computed over $C_s$: patches in its lowest residual quintile are retained, excluding a three-cell neighborhood of its highest residual quintile. If this intersection is empty, the lowest-residual candidate patches provide a fallback; annotated and padded patches remain excluded.

The retained features are pooled into image-level and patch-chunk records. The background bank $B_{bg}$ has 360 records, repeating valid pooled records when necessary. Its mean $\bar b$ summarizes support-specific layout context for query selection, while individual records remain available for \mbox{region classification}.

\subsection{Dual-Evidence Query Selection}
Query selection directs the decoder toward locations that resemble the support devices while discounting similarity to support context. After trainable feature processing, let $x_i$ denote a query token in the selector embedding space and $\bar f=K^{-1}\sum_k f_k$ the foreground mean; hats denote $\ell_2$ normalization. The token ranking is
\begin{equation}
 a_i^{fg}=\langle\widehat{x}_i,\widehat{\bar f}\rangle,
 \quad a_i^{bg}=\langle\widehat{x}_i,\widehat{\bar b}\rangle,
 \quad s_i=a_i^{fg}-\lambda_{\mathrm{sel}}a_i^{bg}.
\label{eq:selector}
\end{equation}
Up to 600 highest-ranked valid tokens initialize a two-layer decoder, which predicts boxes from their spatial references. This allocates the query budget according to support relevance. A shared residual MLP adjusts selector query features with scale 0.1; $\lambda_{\mathrm{sel}}=0.15$.

During adaptation, a selector loss uses feature-token centers inside selected boxes as positives and other valid centers as negatives. Sets $\mathcal P$ and $\mathcal N$ contain up to 128 lowest-scoring positives and 512 highest-scoring negatives, respectively:
\begin{equation}
 \mathcal L_{\mathrm{sel}}=
 \frac{1}{|\mathcal P||\mathcal N|}
 \sum_{i\in\mathcal P,j\in\mathcal N}[0.2-s_i+s_j]_+ .
\label{eq:selectorloss}
\end{equation}

\subsection{Region Evidence and Localization Quality}
Region verification compares each decoded candidate with both evidence banks. RoIAlign~\cite{he2017mask} extracts a $14\times14$ feature grid at each decoded box. The relation head combines similarity to bank means, similarity to individual records, and learned pairwise relations with weights 0.5, 0.5, and 1.0. It uses 256 hidden dimensions, eight region tokens, and the three best-matching support records. Foreground/background logits are normalized jointly:
\begin{equation}
 (p_f(r),p_b(r))=\operatorname{softmax}(z_f(r),z_b(r)).
\label{eq:softmax}
\end{equation}
RoI classification labels are determined by the maximum IoU between each pooled region and annotated targets: regions with IoU $\geq0.5$ are foreground, those with IoU $<0.4$ are background, and the intermediate range is ignored. Multiple RoIs can therefore be positive for one target, while box regression uses one-to-one Hungarian matching~\cite{carion2020detr}.

An IoU-supervised quality head predicts $q_r\in[0,1]$ from region features for localization-aware detection ranking. Binary cross entropy and Smooth L1 anchor predictions to detached IoU targets for assigned or overlapping regions, with zero targets for non-overlapping background regions. A complementary IoU-gap hinge encourages a minimum quality separation between foreground boxes on the same normalized scale. For target $g$, let $\mathcal E_g$ contain ordered pairs assigned to that target by maximum IoU, where $u_i$ is the IoU of region $i$ with $g$, $u_i,u_j\geq0.5$, and $u_i-u_j\geq0.1$. Then
\begin{equation}
 \mathcal L_{\mathrm{rank}}=
 \frac{1}{|\mathcal G'|}\sum_{g\in\mathcal G'}
 \frac{1}{|\mathcal E_g|}\sum_{(i,j)\in\mathcal E_g}
 [(u_i-u_j)-(q_i-q_j)]_+ ,
\label{eq:rank}
\end{equation}
where $\mathcal G'$ includes targets with eligible pairs. Equal target weighting prevents crowded proposal groups from dominating. Box coordinates are detached when constructing targets for the ordering loss.

\subsection{Training Objective and Detection Ranking}
The novel-stage objective combines region cross entropy, box $\ell_1$/GIoU regression, a foreground/background logit margin, selector supervision, and quality losses:
\begin{equation}
\begin{aligned}
\mathcal L={}&2\mathcal L_{\mathrm{CE}}+5\mathcal L_1+2\mathcal L_{\mathrm{GIoU}}
 +0.25\mathcal L_{\mathrm{margin}}\\
 &+0.05\mathcal L_{\mathrm{sel}}+\mathcal L_q
 +\lambda_r\mathcal L_{\mathrm{rank}}+\mathcal L_{\mathrm{dec}}.
\end{aligned}
\label{eq:loss}
\end{equation}
The margin is 0.2 on foreground/background logits, and cross entropy uses background weight 0.2. Quality supervision assigns weight 0.25 to zero-overlap background RoIs and unit weight otherwise. Intermediate decoder predictions receive corresponding Hungarian classification, margin and regression losses, denoted $\mathcal L_{\mathrm{dec}}$. The final 100-step continuation activates $\lambda_r=0.25$ to refine quality ordering after the initial adaptation.

At inference, candidates satisfy $p_f>0.05$. The score used for NMS and reported confidence is
\begin{equation}
 \begin{aligned}
 s_r^{\mathrm{det}}&=p_f(r)\,q_r^{1+0.25h_r}\,d_r,\\
 h_r&=\mathbf 1[p_f(r)-p_b(r)<-0.05].
 \end{aligned}
\label{eq:score}
\end{equation}
The conditional exponent further downweights regions favoring background according to their predicted localization quality. Here $d_r$ discounts redundant boxes. Within same-class overlap components at IoU 0.6, the highest-$p_f$ box is the owner. A lower-confidence box overlapping its owner by at least 0.6 is multiplied by $1-0.35\operatorname{IoU}(r,r_o)$ when $q_r\leq q_{r_o}+0.02$; otherwise $d_r=1$. NMS uses IoU 0.5 and retains at most 100 detections. All categories share these~settings.

\begin{figure}[t]
\centering
\includegraphics[width=\columnwidth]{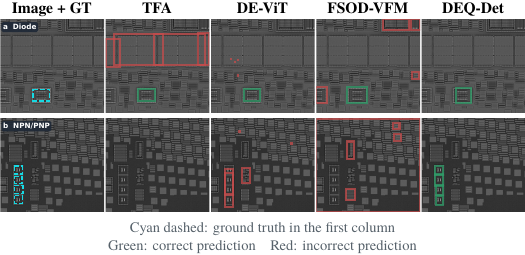}
\caption{Qualitative comparison at 30 shots on Diode and NPN/PNP examples. Columns show ground truth, TFA, DE-ViT, FSOD-VFM, and DEQ-Det. Cyan dashed boxes denote ground truth; green/red boxes denote matched/unmatched predictions at IoU 0.5.}
\label{fig:qualitative}
\end{figure}

\section{Experiments}
\subsection{Data and Evaluation Protocol}
Our dataset comprises image windows cropped from electron micrographs of integrated circuits. It contains 907 training and 282 test windows (3,118/760 instances) from 329/94 source images, respectively. All windows from the same source image remain in a single partition. Each of five fixed splits holds out one novel category---Capacitor, Diode, MOS, NPN/PNP, or Resistor---with the remaining categories used for base training. The novel test subsets contain 35/40/29/31/90 windows and 44/46/55/88/152 target instances, respectively. Their union has 195 distinct windows from 76 test source images; category subsets can overlap in images. All compared methods use these same test manifests.

One shot is one selected novel box, with at most one selected box per support image. At 10 and 30 shots, the novel training data and support banks use exactly those $K$ boxes; locations of other same-class instances are not supplied as ignore regions. All compared methods share the selected support instances. Base/refinement initialization is fixed within each held-out-category split.

Following PASCAL-style few-shot detection~\cite{wang2020tfa,sun2021fsce}, our primary metric is novel-category AP50, computed with COCO evaluation~\cite{lin2014coco}. We additionally report AP75 for the quality ablation. For each seed, we average the five category scores equally, then compute the mean and sample standard deviation over seeds 1/2/3. Support sampling and training randomness are paired; all results use fixed final checkpoints.

\subsection{Implementation and Baselines}
The shared encoder includes a frozen rank-4 LoRA adapter ($\alpha=8$) on query/value projections in its last six blocks~\cite{hu2022lora}. The adapter is pretrained for image reconstruction without device labels and shared across splits and IC-adapted baselines. We retain each split's initialization from 8,000 base-training and 2,000 base-refinement steps. Novel adaptation uses AdamW, batch size two, learning rate $10^{-5}$, and cosine decay for 1,200 steps with 20 warmup steps. A fresh optimizer runs 100 further steps at the same initial learning rate without warmup, enabling Eq.~\eqref{eq:rank}. Adaptation and inference settings are common across categories and support budgets. Images are resized to a maximum side of 1,024 pixels, normalized with ImageNet statistics, and padded to multiples of 14. Training augmentation uses horizontal flips and right-angle rotations. Training uses two RTX 4090 GPUs. At 30 shots, inference with batch size one averages 432 ms per window on an RTX 4090, including model preprocessing and postprocessing; data loading and offline bank construction are excluded.

Original and IC-adapted DE-ViT/FSOD-VFM are reported separately. The adapted variants share the frozen IC encoder with DEQ-Det, but retain their detection pipelines. In particular, IC-adapted DE-ViT retains released background prototypes, and FSOD-VFM retains UPN and SAM2-L. Backbones, pretraining and optimization budgets differ across systems; the comparison evaluates complete detection systems under shared support and test sets.

\begin{table}[t]
\centering
\caption{Five-category macro AP50 at 10 and 30 shots, mean$\pm$sample standard deviation over seeds 1/2/3.}
\label{tab:main}
\small
\renewcommand{\arraystretch}{1.20}
\setlength{\tabcolsep}{3.5pt}
\begin{tabular*}{\columnwidth}{@{\extracolsep{\fill}}lrr@{}}
\toprule
Method & 10-shot & 30-shot \\
\midrule
Template matching & 13.20$\pm$2.00 & 17.17$\pm$2.75 \\
DINOv2 prototype & 12.92$\pm$1.43 & 20.80$\pm$3.28 \\
TFA~\cite{wang2020tfa} & 18.71$\pm$1.56 & 21.93$\pm$1.08 \\
Meta R-CNN~\cite{yan2019meta} & 31.02$\pm$4.81 & \underline{43.32$\pm$5.78} \\
FSCE~\cite{sun2021fsce} & 32.38$\pm$2.17 & 38.65$\pm$0.35 \\
VFA~\cite{han2023vfa} & 32.89$\pm$1.98 & 38.01$\pm$2.18 \\
FPD~\cite{wang2024fpd} & 26.38$\pm$0.93 & 31.74$\pm$3.34 \\
DE-ViT~\cite{zhang2025devit} & 25.91$\pm$1.90 & 30.26$\pm$1.70 \\
DE-ViT + IC adapter & 23.90$\pm$0.60 & 28.98$\pm$5.44 \\
FSOD-VFM~\cite{feng2026fsodvfm} & 31.97$\pm$1.50 & 34.82$\pm$0.07 \\
FSOD-VFM + IC adapter & \underline{35.44$\pm$0.88} & 37.88$\pm$0.43 \\
\midrule \textbf{DEQ-Det} & \textbf{\boldmath 39.41$\pm$2.28} & \textbf{\boldmath 53.03$\pm$4.81} \\
 
\bottomrule
\end{tabular*}
\par\vspace{2pt}\scriptsize
\textbf{Bold} and \underline{underlined} results indicate the best and second-best mean AP50 in each column, respectively. R101: TFA, Meta R-CNN, FSCE, VFA, and FPD; ViT-L/14: the remaining learned methods. IC denotes the frozen IC reconstruction adapter; FSOD-VFM also uses UPN and SAM2-L.
\end{table}

\subsection{Main Results}
DEQ-Det achieves macro AP50 of 39.41$\pm$2.28 and 53.03$\pm$4.81 at 10 and 30 shots, respectively (Table~\ref{tab:main}). The gains over the strongest listed baselines, IC-adapted FSOD-VFM and Meta R-CNN, are \TenShotGain{} and \ThirtyShotGain{} points, respectively. It also leads the IC-adapted foundation-model baselines sharing its frozen encoder. At 30 shots, category AP50 is 62.59/56.08/41.49/64.99/40.03 for Capacitor, Diode, MOS, NPN/PNP, and Resistor, respectively. All five category means improve from 10 to 30 shots. Figure~\ref{fig:qualitative} compares detections on shared Diode and NPN/PNP test images at 30 shots.

\begin{figure}[t]
\centering
\includegraphics[width=\columnwidth]{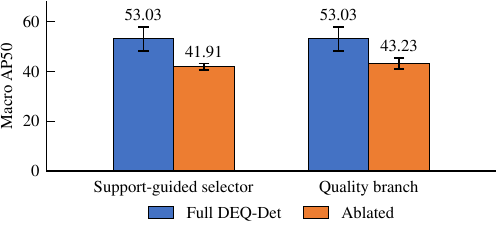}
\caption{Matched 30-shot component ablations: five-category macro AP50, mean and sample standard deviation over three seeds. Blue/orange bars denote full/ablated models.}
\label{fig:ablation}
\end{figure}

\subsection{Component Ablations}
All component ablations use paired support sets and seeds with identical initialization and training schedules. The selector control replaces support-guided selection with feature-norm ranking and removes selector supervision, while retaining RoI evidence and quality scoring. The quality control removes quality supervision and quality-dependent scoring in the same intervention.

Figure~\ref{fig:ablation} evaluates the contributions of query selection and detection ranking. Support-guided selection provides an 11.12-point macro AP50 gain over its matched control. Quality supervision together with quality-dependent scoring adds 9.80 points. Both gains are positive for all three paired seeds. These controls support allocating a fixed query budget to support-relevant candidates and ranking decoded regions with localization-aware confidence. The same quality configuration also improves AP75 by 9.59 points over its matched control.

\textit{Background evidence at comparable recall.} To assess interference suppression, we zero the RoI support-background features during novel adaptation and inference, while retaining selector background evidence ($\lambda_{\mathrm{sel}}=0.15$). Using fixed final checkpoints and score-ordered one-to-one matching at IoU $\geq0.5$, we pool categories within each seed and use each arm's highest threshold reaching the prespecified recall targets of 50/60/70/80/85\%. Across these comparable-recall points, retaining RoI background reduces false positives pooled over seeds by 3.5--9.8\%. At 60\% recall, false positives decrease from 1,004 to 906, with unchanged matched-target counts and reductions for all three paired seeds. RoI background references therefore improve device--background discrimination and reduce spurious candidates for targeted layout inspection. The same control also yields a 0.86-point mean AP50~gain.

\section{Conclusion}
\ConclusionResults

\clearpage
\balance
\bibliographystyle{IEEEtran}
\bibliography{reference,reference_iscas2027_additions}
\end{document}